\documentclass[runningheads]{llncs}
\usepackage{multirow}
\usepackage{graphicx}
\usepackage{caption}
\usepackage{subcaption}
\usepackage{amsmath}
\begin{document}
\title{NOWJ at COLIEE 2023 - Multi-Task and Ensemble Approaches in Legal Information Processing}

\titlerunning{NOWJ at COLIEE 2023}

\author{Thi-Hai-Yen Vuong\inst{1} \and
Hai-Long Nguyen\inst{1} \and
Tan-Minh Nguyen\inst{1} \and \\
Hoang-Trung Nguyen\inst{1} \and
Thai-Binh Nguyen\inst{1} \and
Ha-Thanh Nguyen\inst{2}
}
\authorrunning{Vuong et al.}
%
\institute{University of Engineering and Technology, VNU, Hanoi, Vietnam \email{\{yenvth,long.nh,20020081,20020083,20020328\}@vnu.edu.vn} \and
National Institute of Informatics, Tokyo, Japan \email{nguyenhathanh@nii.ac.jp}
}

\maketitle              
\begin{abstract}
  This paper presents the NOWJ team's approach to the COLIEE 2023 Competition, which focuses on advancing legal information processing techniques and applying them to real-world legal scenarios. Our team tackles the four tasks in the competition, which involve legal case retrieval, legal case entailment, statute law retrieval, and legal textual entailment. We employ state-of-the-art machine learning models and innovative approaches, such as BERT, Longformer, BM25-ranking algorithm, and multi-task learning models. Although our team did not achieve state-of-the-art results, our findings provide valuable insights and pave the way for future improvements in legal information processing.

\keywords{Legal information processing, COLIEE, NOWJ, Multi-task, Ensemble}
\end{abstract}
\section{Introduction}
COLIEE (Competition on Legal Information Extraction/Entailment) is an annual event focusing on advancing legal information processing techniques and applying them to real-world legal scenarios. The competition consists of four tasks, divided into two main categories: case law and statute law competitions. The tasks involve various challenges, such as legal case retrieval, legal case entailment, statute law retrieval, and legal textual entailment.
Fortunately, these challenges can be identified and addressed by referring to previous research in the field.


Chalkidis Ilias and Kampas Dimitrios \cite{chalkidis2019deep} conducted a survey on the early adaptation of Deep Learning in legal analytics, focusing on three main fields: text classification, information extraction, and information retrieval. Their study emphasized the importance of semantic feature representations, which are crucial for the successful application of deep learning in natural language processing tasks. In addition to their analysis, they provided pre-trained legal word embeddings using the WORD2VEC model, which were trained on large corpora containing legislations from various countries, including the UK, EU, Canada, Australia, USA, and Japan.

Nguyen et al. \cite{nguyen2022attentive} addressed the challenge of representing legal documents for statute law document retrieval. They proposed a general approach using deep neural networks with attention mechanisms and developed two hierarchical architectures with sparse attention, named Attentive CNN and Paraformer. These methods were evaluated on datasets in English, Japanese, and Vietnamese. The results showed that attentive neural methods significantly outperformed non-neural methods in retrieval performance across  datasets and languages. Pretrained transformer-based models \cite{vaswani2017attention} achieved better accuracy on small datasets but with high computational complexity, while the lighter weight Attentive CNN performed better on large datasets. The proposed Paraformer outperformed state-of-the-art methods on the 2021 COLIEE dataset, achieving the highest recall and F2 scores in the top-N retrieval task 3.

Vuong et al.'s \cite{vuong2022sm} addressed the challenges of case law retrieval, a complex task involving legal case retrieval and legal case entailment. The difficulties stem from the long length of query and candidate cases, the need to identify legal relations beyond lexical or topical relevance, and the effort required to build large and accurate legal case datasets. 
To tackle these challenges, they proposed a novel approach called the supporting model, which is based on the case-case supporting relation, paragraph-paragraph matching, and decision-paragraph matching strategies. Furthermore, they introduced a method to automatically create a large weak-labeling dataset to overcome the lack of data for training deep retrieval models. Experimental results demonstrated that their solution achieved state-of-the-art results for both case retrieval and case entailment~phases.

As the first time participating COLIEE, the NOWJ team aims to tackle the 2023 competition by utilizing state-of-the-art machine learning models and innovative approaches. For Tasks 1 and 2, we rely on BERT and Longformer pre-training models without using any external data. Additionally, Task 2 employs an internal data generation method based on Vuong et al \cite{vuong2022sm} method to overcome the lack of data and enhance the legal case retrieval process. For Task 3, our team utilizes a two-phase retrieval system that employs the BM25-ranking algorithm and a BERT-based model for re-ranking, along with additional techniques to handle relations between articles and exploit information from the Legal Textual Entailment Data Corpus tasks. Lastly, for Task 4, we use a multi-task model with a pre-trained Multilingual BERT as the backbone.

Although we did not achieve state-of-the-art results, our findings in this competition provide good insights and pave the way for further improvements in legal information processing. 

\section{Task Introduction}
The competition consists of four tasks focusing on case law and statute law. Tasks 1 and 2 address the case law challenges, while tasks 3 and 4 target the statute law problems. Task 1, the legal case retrieval task, requires participating systems to read a new case and extract supporting cases from a case law corpus. Task 2, the legal case entailment task, involves the systems identifying a paragraph from existing cases that entails the decision of a new case. Task 3, part of the statute law competition, requires systems to retrieve relevant Japanese civil code statutes. Finally, Task 4 involves the systems confirming the entailment of a yes/no answer from the retrieved civil code statutes. The competition aims to advance legal information processing techniques and apply them to real-world legal scenarios through the development of innovative and efficient systems.

\subsection{Task 1\&2: The Legal Case Retrieval and Entailment Task}
In the legal case retrieval task, participating systems are required to examine a new legal case $Q$ and extract supporting cases $S_1, S_2, ... S_n$ from a case law corpus to provide backing for the decision of $Q$.
In real-world legal scenarios, lawyers and legal professionals often need to discover pertinent case laws to substantiate their arguments in court. This task emulates the process of identifying precedents and offering legal arguments based on previous cases' analysis. Automating this procedure with an efficient system allows saving time and resources while ensuring accuracy and relevance in supporting cases.

The UA team \cite{rabelo2023semantic} developed a successful approach for this task in the COLIEE 2022 competition. Their method combined semantic similarity representation at the sentence level and a Gradient Boosting binary classifier trained on 10-bin histograms containing similarity scores between sentences of the query and candidate cases. Additionally, they applied simple pre- and post-processing heuristics to generate the final results. As a result, they achieved the highest ranking among all competitors, outperforming 9 teams with a total of 26 submissions. 

For the legal case entailment task, participating systems must pinpoint a specific paragraph from existing cases that involves the decision of a new legal case $Q$. Given a decision $Q$ for a new case and its relevant counterpart $R$, the systems must identify which paragraph supports Q's decision.
In actual practice, it is crucial for lawyers and other professionals in law to locate precise arguments or reasoning within existing situations while building strong foundations for their cases. Utilizing an efficient system can help save time, guarantee accuracy when automating this procedure, strengthen their arguments by detecting relevant paragraphs from existing ones.

The NM team \cite{rosa2022billions} conducted experiments with zero-shot models in the legal case entailment task. Utilizing large language models, such as GPT-3, they found that scaling the number of parameters in a language model improved the F1 score by more than 6 points compared to their previous zero-shot result. This suggests that larger models may possess stronger zero-shot capabilities, at least for this specific task. Their 3B-parameter zero-shot model outperformed all other models, including ensembles, in the COLIEE 2021 test set and achieved the best performance of a single model in the COLIEE 2022 competition, ranking second only to an ensemble consisting of the 3B model and a smaller version of the same model. Despite the challenges posed by large language models, particularly latency constraints in real-time applications, the NM team demonstrated the practical use of their zero-shot monoT5-3b model in a search engine for legal documents.

\subsection{Task 3\&4: The Statute Law Retrieval and Entailment Task}

For statute law retrieval tasks involving answering yes/no questions about the Japanese Civil Code statutes $(S_1, S_2, ... , S_n)$ participating systems will read a question $(Q)$ then retrieve relevant data from databases accordingly. This simulates locating specific provisions needed when determining if certain laws apply given various situations encountered during actual practice sessions among lawyers working with clients. This work seeks advice regarding potential outcomes resulting from particular instances requiring clarification between parties involved within disputes occurring throughout professional settings worldwide today where many people interact daily exchanging ideas concerning different aspects life experience overall.
In actual practice, lawyers and legal professionals must quickly identify the relevant statutes governing specific situations. Automating this process with an efficient system can save time and ensure accuracy when interpreting and applying laws to various cases.

The HUKB team \cite{yoshioka2023hukb} proposed a method that utilized three different Information Retrieval (IR) systems. Their new IR system was designed to measure the similarity of descriptions of judicial decisions between questions and articles. In addition to this new system, they also employed an ordinal keyword-based IR system (BM25) and a BERT-based IR system that was proposed in COLIEE 2020. Due to the diverse characteristics of these systems, ensembled results provided better recall without sacrificing much precision. The HUKB group's ensemble, which combined their newly proposed IR system and the keyword-based IR system, achieved the best performance for COLIEE 2022 Task 3. 

For the statute law entailment task, participating systems are responsible for confirming whether Japanese Civil Code provisions entail a yes/no answer to question $Q$. Given $Q$ along with relevant statutes $(S_1, S_2, ... , S_n)$, systems must determine if they support a "YES" ("Q") or "NO" ("not Q") response.

In real-world legal scenarios, it is essential for lawyers and legal professionals to assess the implications of legal provisions on specific cases accurately. Automating this process with an efficient system can help save time, ensure accuracy, and enhance decision-making by effectively evaluating the relevance and implications of these provisions.

The KIS team \cite{fujita2023legal} developed a successful approach for COLIEE 2022 Task 4, which aimed to solve the textual entailment part of the Japanese legal bar examination problems. By employing an ensemble of a rule-based method and a BERT-based method, and utilizing data augmentation and modular ensembling techniques, they improved the correct answer ratio. Their approach integrated additional proposed methods, such as Sentence-BERT for data selection and person name inference for replacing anonymized symbols. As a result, the KIS team achieved the best score among the Task 4 submissions with a correct answer ratio of 0.6789 in accuracy on the formal run test dataset.

\section{Methodology}

\subsection{Task 1-2: The Legal Case Retrieval and Entailment Task}

The relationship between base case - case in the legal case retrieval task  and decision - paragraph in the legal case entailment task is relatively similar, the candidate case supports the base case or the paragraph supports the decision. The length of the legal case, paragraphs, and decision makes a significant distinction between these two tasks. Due to the extremely extensive texts in both the query and candidate instances, the work of legal case retrieval is quite difficult. To tackle these challenges, Vuong et al. proposed a novel approach called the supporting model, which is based on the case-case supporting relation, paragraph-paragraph matching, and decision-paragraph matching strategies \cite{vuong2022sm}.

On the other hand, the relationship between the paragraphs in the legal case will be lost if only matching at the paragraph/decision - paragraph level. Therefore, we build a case-level matching model to evaluate the support relationship of the base case and candidate cases. By incorporating both local and global attention mechanisms, Longformer is proposed to effectively encode long texts with thousands of tokens, overcome inability to process long documents limitations that other pre-trained language models have \cite{beltagy2020longformer}. Longformer could capture effectively the legal case documents with the characteristic of long length.

In this study, we build two matching phases to solve the legal case retrieval and entailment task: mono tatching (paragraph/decision level) and panorama matching (case level).

\subsubsection{Pre-processing}

Firstly, we implemented some simple pre-processing steps:

\begin{itemize}
    \item Due to the majority of queries being in English, we have chosen to eliminate all French content from the data, even if the legal case that involves or includes French translations.
    \item Segment case into paragraphs based on common struture of the legal case.
    \item Extraction of case's year through a rule-based method. Our assumption is that noticed cases could not be more recent than the base case. Thus, these years are used to filter out candidate cases that include dates more recent than the most recent date mentioned in a base case.
    \item Removal of redundant characters using regex. We removed duplicate endline, space characters and punctuations (exclude period, commas, question marks, etc.).
    \item  Detect important passages by using heuristic. The placement of a paragraph in the legal case document also reveals its importance level such as whether paragraph is in "I. Introduction" or "II. Background", and so on. Beside, some of the paragraphs in base case quote other cases, in where placeholders like "SUPPRESSED" are used of the cited cases, these  paragraphs also contain important content and words in comparison with other case law.
    
\end{itemize}

\subsubsection{Mono Matching - Paragraph/Fragment level}
This phase is a combination of two models which perform lexical and semantic matching, respectively:

\begin{itemize}
    \item Lexical matching: using BM25 calculates the relevance score betwent paragraph/decision - paragraph based on the frequency of query terms in the query and their frequency in the entire collection of documents, as well as other factors such as document length and average query length.
    \item Semantic matching: to extract the semantic relationship we built a supporting model \cite{vuong2022sm}. It is fine-tuned with the weak label dataset and the legal case entailment dataset.
\end{itemize}

Particularly for the legal case retrieval task, the search space is huge with 4400 candidates case in the train set and 1300 candidates cases in the test set. To ensure the model's performance, we initially narrow down the search space by utilizing a lexical model to select potential candidates. For each paragraph of a base case, we retrieved the top 200 candidate paragraphs. We identify candidate cases through the returned candidate paragraphs. If a candidate case appears more than twice, we will retain the highest score. Ultimately, we will retain the cases with the best scores, up to a maximum of $k$ cases.

\subsubsection{Panorama Matching - Case level}
In this stage, a Longformer model was implemented to compare a base case with candidate cases based on their similarities and relatedness in the panorama. In consideration of the typical average length of legal cases, which is approximately 3000 tokens, it was observed that a (base, candidate) case pair would exceed the token limitation of the Longformer model. To overcome this limitation, it was deemed necessary to curate the input by retaining only the most important paragraphs of the base case. This allowed us to mitigate the input length while still preserving the salient information necessary for effective matching with relevant candidate cases. This curation approach ensured that the Longformer model could successfully process the input data, thereby achieving superior matching results while also enhancing the efficiency of the matching process. 

In the process of constructing the training dataset, we initially selected cases that were labeled as noticed to the query and assigned them as positive samples. We then proceeded to identify negative samples by considering cases that were labeled as not noticed for each query case, but were retrieved by the mono matching phase. As a result, the resulting dataset consisted of pairs of cases (base, candidate) with a label ratio of 1:2 for positive and negative samples, respectively.

\subsection{Task 3-4: The Statue Law Retrieval and The Legal Textual Entailment Data Corpus}
Due to the good performance on the recall score of the BM25 model on the retrieval task, which is proven on previous works \cite{kim2022legal} \cite{rosa2021yes} \cite{shao2020bert}, the retrieval problem of task 3 is tackled using two-stage ranking including BM25 ranking model as first stage and BERT-base ranking model as the second stage. The BERT-based ranking model utilizes multi-task learning, which combines the goal of the retrieval (task 3) and textual entailment (task 4) problems.

\subsubsection{First-stage BM25 ranking}
The BM25 model \cite{robertson1995okapi} is a probabilistic relevance-based model that is widely used in the retrieval field. As it primarily operates on statistical processes and computes the relevance score through a single mathematical formula, it has a fast retrieval time in large corpora. The BM25 model is employed in this work to narrow down the candidate articles for the training phase. Additionally, its relevance score is utilized to improve the recall score during the testing phase by ensembling it with the ranking-BERT's relevance score.


According to the experiment, limiting the negative samples by BM25 ranking significantly reduces the training time but still remains the training efficency. Since the BM25 ranking selects the articles that are relevant to the given query in the lexical level, restricting the candidate documents by the top result of BM25 could make the re-ranking model has ability of distinguishing lexical-related and semantic-related candidates. To conclude, employing the pre-ranking BM25 model can make the ranking model more effectively in both training and inference phase.

\subsubsection{Multi-task model with BERT}
Although BM25 model can achieve a good recall score with a sufficient number of top candidates, its precision score is still very low. Therefore, a re-ranking BERT-based model is utilized for filtering the BM25's candidates and improving the precision score while keeping the recall score consistent. In this work, the re-ranking models'architecture is determined based on the fact that whether a query is considered as \textit{yes} or \textit{no} highly depends on the perspective of the legal experts. This means that when a query is labeled as \textit{yes} (or \textit{no}) by the legal experts, they tend to find the relevant articles that support their perspective. In other words, the relevant candidates are highly correlated with the \textit{yes/no} result. Arcording the above observation, a multi-task model which has two output heads, including the retrieval head and textual entailment head, is employed as the re-ranking model. The model utilized the BERT architecture as the backbone and two output heads were added on top of it. 

For conveniently comparing the performance of model when working with the different languages, the \textbf{bert-base-multilingual-uncased} pre-train parameters are used as the initialization for the model's back-bone. 

\subsection{Ensemble Model}

Integrating multiple models can lead to a more efficient and effective solution, as it enables us to leverage the individual strengths of each model.  Consequently, we establish emsemble methods  which handle the intricacies and complexities that may not be addressed by a single model, while also enhancing the overall performance and precision of the system. 

\subsubsection{Boosting Ensemble}
Boosting ensembles can be utilized in the retrieval process to filter out negative samples step-by-step. In each boosting step, a subset of candidates will be eliminated by a ranking model. This approach could combine the advantages of multiple models, which leads to an improvement in the accuracy and efficiency of the retrieval system. For example, with each input query, the retrieval system has to consider $n$ legal documents in the database denoted as a set $S = \{s_1, s_2, ..., s_n\}$. After first boosting step, a subset of $S$ will be removed from the potential candidate set by a ranking model, which forms a new candidate set with $m$ elements denoted as $S_1 = \{s_{i_1}, s_{i_2}, ..., s_{i_m}\}$ where $i_j \in \{1, 2, 3, ..., n\}$ and $m < n$. The process is performed similarly in the remaining boosting step. In this investigation, ranking models from the lexical level to the semantic level are used in the ranking phases, respectively.

\subsubsection{Weighted Ensemble}

According to our experiments, the Lexical Matching Models often gain a relatively good recall score, while the Semantic Matching Models improve the precision score. So, combining the results of these two models could help to raise the overall F2 score. The relevance score of the semantic matching model is ensembled with the lexical matching model using the equation \ref{equation:rel_score}.
\begin{equation}
    relevant\_score = \alpha * bm25\_score + \beta * bert\_score
    \label{equation:rel_score}
\end{equation}
The $bm25\_score$ is the relevance score given by the BM25 model (or Lexical Matching Model), $bert\_score$ is the relevance score given by the re-ranking BERT-based model (or Semantic Matching Model). The min-max normalization is performed with both $bm25\_score$, $bert\_score$, and the $relevant\_score$. The min-max normalization is computed as in Equation \ref{equation:min-max norm}.
\begin{equation}
    x'=\frac{x-min(x)}{max(x)-min(x)}
    \label{equation:min-max norm}
\end{equation}
where $x$ is an original value, $x'$ is the normalized value.
The trail-threshold inference strategy which determines the relevant articles based on the highest relevance score is used. A candidate is considered as relevant article if its relevance score satisfies the Equation \ref{equation:trail_threshold}. 
\begin{equation}
    \frac{highest\_score - candidate\_score}{highest\_score} <= trail\_threshold
    \label{equation:trail_threshold}
\end{equation}
The $highest\_score$ is the highest relevance score among all candidates of the given query and the $candidate\_score$ is the relvant score of the considering candidate. All the relevance scores here are the score after the combination and normalization process. For tuning optimal $\alpha$, $\beta$ and $trail\_threshold$, a grid-search process is conducted on the development set.

\subsubsection{Voting ensemble}
In many cases, the voting ensemble method, which uses the predicted results of several models and determines the final label based on the majority, could be an effective error-correcting process \cite{dietterich2000ensemble}. This approach could significantly improve the overall results by considering the perspectives of multiple models. In this study, the voting ensemble method was employed for one of the submissions, which turned out to be the best run among all runs.

\section{Experiment and Result}
\subsection{Measuring}
The evaluation metrics of Tasks 1 and 2 are precision, recall, and F-measure. All the metrics are micro-average, which means the evaluation measure is calculated using the results of all queries. The definition of these measures is as follows:
\begin{equation}
    \text{Precision} = \frac{\text{\# correctly retrieved cases (paragraphs) for all queries}}{\text{\# retrieved cases (paragraphs) for all queries}}
\end{equation}
\begin{equation}
    \text{Recall} = \frac{\text{\# correctly retrieved cases (paragraphs) for all queries}}{\text{\# relevant cases (paragraphs) for all queries}}
\end{equation}
\begin{equation}
    \text{F-measure} = \frac{2\times\text{Precision}\times\text{Recall}}{\text{Precision}+\text{Recall}}
\end{equation}

For Task 3, evaluation measures are precision, recall and F2-measure. All the metrics are macro-average (evaluation measure is calculated for each query and their average is used as the final evaluation measure) instead of micro-average (evaluation measure is calculated using results of all queries). The definition of these measures is as follows:
\begin{equation}
    \text{Precision} = \text{average of}\frac{\text{\# correctly retrieved articles for each query}}{\text{\# retrieved articles for each query}}
\end{equation}
\begin{equation}
    \text{Recall} = \text{average of}\frac{\text{\# correctly retrieved articles for each query}}{\text{\# relevant articles for each query}}
\end{equation}
\begin{equation}
    \text{F-measure} = \text{average of}\frac{5\times\text{Precision}\times\text{Recall}}{4\times\text{Precision}+\text{Recall}}
\end{equation}

For Task 4, the evaluation measure will be accuracy, with respect to whether the yes/no question was correctly confirmed:
\begin{equation}
    \text{Accuracy} = \frac{\text{\# queries which were correctly confirmed as true or false}}{\text{\# all queries}}
\end{equation}

\subsection{Task 1}
An official corpus has been provided for the evaluation of legal case retrieval models in COLIEE-2023 Task 1. The corpus relates to a database of mainly Federal Court of Canada case laws from Compass Law. Table \ref{tab: data analyse task1} outlines the dataset statistics for Task 1, which includes 959 query cases against 4400 candidate cases in the training set and 319 query cases against 1335 candidate cases in the test dataset. The average number of paragraphs per case in the training dataset is 42.29, while the testing dataset has an average of 37.51 paragraphs per case. Further analysis revealed an average of 4.47 noticed cases in the training dataset and 2.69 noticed cases in the testing dataset.
\begin{table}[ht]
\caption{Statistics of the dataset for Task 1} 
\label{tab: data analyse task1}
\centering
\begin{tabular}{|lcc|}
\hline
\multicolumn{1}{|l|}{\textbf{}}                      & \multicolumn{1}{c|}{\textbf{Train}} & \textbf{Test} \\ \hline
\multicolumn{1}{|l|}{\textbf{\# queries}}         & \multicolumn{1}{c|}{959}            & 319           \\ \hline
\multicolumn{1}{|l|}{\textbf{\# candidate cases}} & \multicolumn{1}{c|}{4400}           & 1335          \\ \hline
\multicolumn{3}{|l|}{\textbf{\# noticed cases per query}}         \\ \hline
\multicolumn{1}{|l|}{Min}     & \multicolumn{1}{c|}{1}      & 1      \\ \hline
\multicolumn{1}{|l|}{Max}     & \multicolumn{1}{c|}{34}     & 17      \\ \hline
\multicolumn{1}{|l|}{Average} & \multicolumn{1}{c|}{4.67}   & 2.69      \\ \hline
\multicolumn{3}{|l|}{\textbf{\# paragraphs per case}}             \\ \hline
\multicolumn{1}{|l|}{Min}     & \multicolumn{1}{c|}{2}      & 3      \\ \hline
\multicolumn{1}{|l|}{Max}     & \multicolumn{1}{c|}{1117}   & 617    \\ \hline
\multicolumn{1}{|l|}{Average} & \multicolumn{1}{c|}{42.29}  & 37.51  \\ \hline
\end{tabular}
\end{table}

We implement the lexical matching based on BM25 method using Elasticsearch \footnote{https://www.elastic.co/}. We experimented five options of top-$k$ candidate cases: $N=10, N=20, N=50, N=100, N=200$ with different features. $Recall@k$ measure is used for evaluating the list of returned candidates. Recall@k is (Number of correctly predicted articles in the $top-k$ results) / (Total number of gold articles). Table \ref{tab: top-n elasticsearch} presents experimental results of the lexical matching method. In $k=10$ and $k=20$, using only important paragraphs instead of a whole case improved Elasticsearch performance significantly. Specifically, there is a $77\%$ increase in Recall@$k$ for $k=10$ and a $32\%$ increase for $k=20$. Additionally, including Year into queries also showed an improvement in Elasticsearch performance.
Based on this evaluation, we chose top 200 candidates from Elasticsearch method.
\begin{table}[ht]
\caption{Top-$k$ recall score of the lexical matching method}
\centering
\begin{tabular}{|l|l|l|l|l|l|}
\hline
\multicolumn{1}{|c|}{\textbf{Top $k$}} &
  \multicolumn{1}{c|}{\textbf{10}} &
  \multicolumn{1}{c|}{\textbf{20}} &
  \multicolumn{1}{c|}{\textbf{50}} &
  \multicolumn{1}{c|}{\textbf{100}} &
  \multicolumn{1}{c|}{\textbf{200}} \\ \hline
  All paragraphs & 0.1783 & 0.2961 & 0.4803 & 0.5997 & 0.7190 \\ \hline
  Important paragraphs & 0.3076 & 0.3921 & 0.5176 & 0.6161 & 0.7257 \\ \hline
  All paragraphs + Year & 0.1877 & 0.3110 & 0.5026 & 0.6100 & 0.7152 \\ \hline
  Important paragraphs + Year & 0.3334 & 0.4211 & 0.5319 & 0.6275 & 0.7290 \\ \hline
\end{tabular}
\label{tab: top-n elasticsearch}
\end{table}

Our team has submited 3 runs for the private test. The first run named \textbf{NOWJ.bestsingle} employed the mono matching model, which used the boosting ensemble method to eliminate candidates judged as non-potential and retained at most only one candidate for each paragraph in the base case. The second one named \textbf{NOWJ.ensemble} used the panorama matching model. The final run is \textbf{NOWJ.d-ensemble} used the voting ensemble of results from the two previous runs. 

Table \ref{tab: task1 result} illustrates the leaderboard of Task 1, NOWJ is our team. We submitted three official runs and ranked third among all teams. THUIR achieves the best performance in terms of F1 and Recall scores, while UFAM has the highest Precision score. The search space is large, which is 1335 candidate cases against 319 query cases in the testing dataset. Furthermore, the average case length in the training set is 3712.71 words, which prevents models from effectively aggregating information. Further analysis on our results revealed that combining the mono and panorama matching methods improved model's performance. 

\begin{table}[ht]
\caption{Results on the test set of Task 1}
\centering
\begin{tabular}{|lllccc|}
\hline
\multicolumn{1}{|c|}{\textbf{\#}} & \multicolumn{1}{c|}{\textbf{Team}} & \multicolumn{1}{c|}{\textbf{Run}}                  & \multicolumn{1}{c|}{\textbf{F1}} & \multicolumn{1}{c|}{\textbf{P}} & \textbf{R}                  \\ \hline
\multicolumn{6}{|l|}{\textbf{Other team's best results}}                                                                                                                                                                         \\ \hline
\multicolumn{1}{|l|}{1}             & \multicolumn{1}{l|}{THUIR}         & \multicolumn{1}{l|}{thuirrun2}                 & \multicolumn{1}{c|}{0.3001}      & \multicolumn{1}{c|}{0.2379}     & 0.4063                      \\ \hline
\multicolumn{1}{|l|}{2}             & \multicolumn{1}{l|}{IITDLI}        & \multicolumn{1}{l|}{iitdli\_task1\_run3}       & \multicolumn{1}{c|}{0.2874}      & \multicolumn{1}{c|}{0.2447}     & 0.3481                      \\ \hline
\multicolumn{1}{|l|}{4}             & \multicolumn{1}{l|}{JNLP}          & \multicolumn{1}{l|}{jnlp\_cl\_3\_dates}       & \multicolumn{1}{c|}{0.2604}      & \multicolumn{1}{c|}{0.2044}     & 0.3586                      \\ \hline
\multicolumn{1}{|l|}{5}             & \multicolumn{1}{l|}{UA}            & \multicolumn{1}{l|}{pp\_0.8\_10\_3}            & \multicolumn{1}{c|}{0.2555}      & \multicolumn{1}{c|}{0.2847}     & 0.2317                      \\ \hline
\multicolumn{1}{|l|}{6}             & \multicolumn{1}{l|}{UFAM}          & \multicolumn{1}{l|}{task1\_2023\_k50t036\_3}  & \multicolumn{1}{c|}{0.2545}      & \multicolumn{1}{c|}{0.2975}     & 0.2224                      \\ \hline
\multicolumn{1}{|l|}{7}             & \multicolumn{1}{l|}{YR}            & \multicolumn{1}{l|}{task1\_yr\_run1}           & \multicolumn{1}{c|}{0.1377}      & \multicolumn{1}{c|}{0.1060}     & 0.1967                      \\ \hline
\multicolumn{1}{|l|}{8}             & \multicolumn{1}{l|}{LLNTU}         & \multicolumn{1}{l|}{task1\_llntucliiss\_2023} & \multicolumn{1}{c|}{0.0000}      & \multicolumn{1}{c|}{0.0000}     & 0.0000                      \\ \hline
\multicolumn{6}{|l|}{\textbf{Our results}}                                                                                                                                                                                       \\ \hline
\multicolumn{1}{|l|}{3}             & \multicolumn{1}{l|}{NOWJ}          & \multicolumn{1}{l|}{nowj.d-ensemble}          & \multicolumn{1}{c|}{0.2757}      & \multicolumn{1}{c|}{0.2263}     & 0.3527                      \\ \hline
\multicolumn{1}{|l|}{3}             & \multicolumn{1}{l|}{NOWJ}          & \multicolumn{1}{l|}{nowj.ensemble}            & \multicolumn{1}{c|}{0.2756}      & \multicolumn{1}{c|}{0.2272}     & 0.3504                      \\ \hline
\multicolumn{1}{|l|}{3}             & \multicolumn{1}{l|}{NOWJ}          & \multicolumn{1}{l|}{nowj.bestsingle}          & \multicolumn{1}{l|}{0.2573}      & \multicolumn{1}{l|}{0.2032}     & \multicolumn{1}{l|}{0.3504} \\ \hline
\end{tabular}
\label{tab: task1 result}
\end{table}

\subsection{Task 2}
Table \ref{tab: data analyse task2} shows the dataset statistics for Task 2, which includes 625 queries in the training dataset and 100 queries in the testing dataset. The average number of paragraphs per query in the training dataset is 35, while the testing dataset has an average of 37.65 paragraphs per query. There is an average of 1.17 entailed fragments in the training dataset and 1.20 entailed fragments in the testing dataset. The average length of a entailed fragment in both dataset is approximately 34 words.
\begin{table}[ht]
\caption{Statistics of the dataset for Task 2}
\label{tab: data analyse task2}
\centering
\begin{tabular}{|lcc|}
\hline
\multicolumn{1}{|l|}{}                          & \multicolumn{1}{c|}{\textbf{Train}} & \textbf{Test}         \\ \hline
\multicolumn{1}{|l|}{\textbf{\# query case}} & \multicolumn{1}{c|}{625}            & 100                   \\ \hline
\multicolumn{3}{|l|}{\textbf{\# paragraphs per query}}                                                     \\ \hline
\multicolumn{1}{|l|}{Min}                       & \multicolumn{1}{c|}{3}              & 3                     \\ \hline
\multicolumn{1}{|l|}{Max}                       & \multicolumn{1}{c|}{283}            & 170                   \\ \hline
\multicolumn{1}{|l|}{Avg}                       & \multicolumn{1}{c|}{35}             & 37.65                 \\ \hline
\multicolumn{3}{|l|}{\textbf{\# entailed fragments per query}}                                             \\ \hline
\multicolumn{1}{|l|}{Min}                       & \multicolumn{1}{c|}{1}              & 1                     \\ \hline
\multicolumn{1}{|l|}{Max}                       & \multicolumn{1}{c|}{5}              & 4                     \\ \hline
\multicolumn{1}{|l|}{Avg}                       & \multicolumn{1}{c|}{1.17}           & 1.20                  \\ \hline
\multicolumn{3}{|l|}{\textbf{\# words per fragments}}                                                      \\ \hline
\multicolumn{1}{|l|}{Min}                       & \multicolumn{1}{c|}{4}              & \multicolumn{1}{c|}{8} \\ \hline
\multicolumn{1}{|l|}{Max}                       & \multicolumn{1}{c|}{115}            & \multicolumn{1}{c|}{111} \\ \hline
\multicolumn{1}{|l|}{Avg}                       & \multicolumn{1}{c|}{34}             & \multicolumn{1}{c|}{33.18} \\ \hline
\end{tabular}
\end{table}

The problem raised on task 2 is the textual entailment of the relevant cases. In this task, we submitted three runs for the private test. The first run, named \textbf{NOWJ.non-empty}, employed the mono matching model using the weighted ensemble method. The second one named \textbf{NOWJ.hp} and the final run (\textbf{NOWJ.hr}) used the boosting ensemble method to eliminate candidates judged as non-potential and then applied the weighted ensemble method to get the final result with different hyperparams setting. 

Table \ref{tab: task2 result} illustrates the leaderboard of Task 2. We submitted three official runs and ranked third among all teams. CAPTAIN achieves the best performance in terms of F1 and Recall scores while THUIR has the highest Precision score. Our team utilized BERT-based models to tackle Task 1 and 2 in COLIEE 2023. Since the introduction of BERT in 2018, newer pre trained language models with improved performance have been developed,  as GPT-3, T5. These large models have more parameters and are trained on larger datasets. However, deploying large language models could be a challenge due to infrastructure and training cost requirements. 
\begin{table}[ht]
\caption{Results on the test set of Task 2}
\label{tab: task2 result}
\centering
\begin{tabular}{|lllccc|}
\hline
\multicolumn{1}{|c|}{\textbf{\#}} & \multicolumn{1}{c|}{\textbf{Team}} & \multicolumn{1}{c|}{\textbf{Run}}                  & \multicolumn{1}{c|}{\textbf{F1}} & \multicolumn{1}{c|}{\textbf{P}} & \textbf{R}                  \\ \hline
\multicolumn{6}{|l|}{\textbf{Other team's best results}}                                                                                                                                                                       \\ \hline
\multicolumn{1}{|l|}{1}           & \multicolumn{1}{l|}{CAPTAIN}       & \multicolumn{1}{l|}{mt5l-ed}                       & \multicolumn{1}{c|}{0.7456}      & \multicolumn{1}{c|}{0.7870}     & 0.7083                      \\ \hline
\multicolumn{1}{|l|}{2}           & \multicolumn{1}{l|}{THUIR}         & \multicolumn{1}{l|}{thuir-monot5}                  & \multicolumn{1}{c|}{0.7182}      & \multicolumn{1}{c|}{0.7900}     & 0.6583                      \\ \hline
\multicolumn{1}{|l|}{3}           & \multicolumn{1}{l|}{JNLP}          & \multicolumn{1}{l|}{jnlp\_bm\_cl\_1\_pr\_1}        & \multicolumn{1}{c|}{0.6818}      & \multicolumn{1}{c|}{0.7500}     & 0.6250                      \\ \hline
\multicolumn{1}{|l|}{4}           & \multicolumn{1}{l|}{IITDLI}        & \multicolumn{1}{l|}{iitdli\_task2\_run2}           & \multicolumn{1}{c|}{0.6727}      & \multicolumn{1}{c|}{0.7400}     & 0.6167                      \\ \hline
\multicolumn{1}{|l|}{5}           & \multicolumn{1}{l|}{UONLP}         & \multicolumn{1}{l|}{task2\_test\_no\_labels\_2023} & \multicolumn{1}{c|}{0.6387}      & \multicolumn{1}{c|}{0.6441}     & 0.6333                      \\ \hline
\multicolumn{1}{|l|}{7}           & \multicolumn{1}{l|}{LLNTU}         & \multicolumn{1}{l|}{task2\_llntukwnic\_2023}       & \multicolumn{1}{c|}{0.1818}      & \multicolumn{1}{c|}{0.2000}     & 0.1667                      \\ \hline
\multicolumn{6}{|l|}{\textbf{Our results}}                                                                                                                                                                                     \\ \hline
\multicolumn{1}{|l|}{6}           & \multicolumn{1}{l|}{NOWJ}          & \multicolumn{1}{l|}{nowj.d-ensemble}               & \multicolumn{1}{c|}{0.2757}      & \multicolumn{1}{c|}{0.2263}     & 0.3527                      \\ \hline
\multicolumn{1}{|l|}{6}           & \multicolumn{1}{l|}{NOWJ}          & \multicolumn{1}{l|}{nowj.ensemble}                 & \multicolumn{1}{c|}{0.2756}      & \multicolumn{1}{c|}{0.2272}     & 0.3504                      \\ \hline
\multicolumn{1}{|l|}{6}           & \multicolumn{1}{l|}{NOWJ}          & \multicolumn{1}{l|}{nowj.bestsingle}               & \multicolumn{1}{l|}{0.2573}      & \multicolumn{1}{l|}{0.2032}     & \multicolumn{1}{l|}{0.3504} \\ \hline

\end{tabular}
\end{table}

\subsection{Task 3}
Firstly, a simple statistical analysis has been conducted for determining the top-k candidates which could be gained from the BM25 model. The table \ref{tab:task3-bm25-score} shows the top-k candidates and the corresponding recall score. Based on that result, the top 30 candidates are used for the training phase for reducing the training time but still remaining a good recall score and the top 500 is used for the inference phase. 

In the re-ranking phase, the pre-train \textbf{bert-base-multilingual-uncased} \footnote{https://huggingface.co/bert-base-multilingual-uncased} is employed as the back-bone and two heads corresponding to two learning tasks are added on top. In this investigation, two models are trained with different languages: English and Japanese in 10 epochs. The first model uses the Japanese data for the training phase and the second model mainly uses the English data. 

The relevance score of re-ranking BERT is combined with the BM25 score for determining the final relevance score arcording to the equation.

For making the final submission, an ensemble process based on voting is conducted to combine the BM25-score, Multi-Task JP and Multi-Task EN model's relevance score, which achieves the third rank among all teams. The results of the ensemble method, along with the best results from other teams, on the 2023 private test are described in the Table \ref{tab: task3 method result}. According to the results, our ensemble model achieves better \textit{Precision} than the best run of the second team (JNLP3) but has lower \textit{Recall}. 

\begin{table}[ht]
\caption{Recall score of corresponding top-$k$}
\centering
\begin{tabular}{|c|c|}
\hline
\textbf{Top-$k$ candidates} & \textbf{Recall score} \\ \hline
30               & {0.7784}       \\ \hline
100              & 0.8513       \\ \hline
200              & 0.8926       \\ \hline
500              & {0.9487}      \\ \hline
\end{tabular}
\label{tab:task3-bm25-score}
\end{table}

\begin{table}[ht]
\caption{Results on the private test of Task 3}
\centering
\begin{tabular}{|lllccc|}
\hline
\multicolumn{1}{|c|}{\textbf{\#}} & \multicolumn{1}{c|}{\textbf{Team}} & \multicolumn{1}{c|}{\textbf{Best Run}} & \multicolumn{1}{c|}{\textbf{F2}} & \multicolumn{1}{c|}{\textbf{P}} & \textbf{R} \\ \hline
\multicolumn{6}{|l|}{\textbf{Other team's result}}                                                                                                                                                \\ \hline
\multicolumn{1}{|l|}{1}           & \multicolumn{1}{l|}{CAPTAIN}       & \multicolumn{1}{l|}{allEnssMissq}      & \multicolumn{1}{c|}{0.7569}      & \multicolumn{1}{c|}{0.7261}     & 0.7921     \\ \hline
\multicolumn{1}{|l|}{2}           & \multicolumn{1}{l|}{JNLP}          & \multicolumn{1}{l|}{JNLP3}             & \multicolumn{1}{c|}{0.7451}      & \multicolumn{1}{c|}{0.6452}     & 0.8218     \\ \hline
\multicolumn{1}{|l|}{4}           & \multicolumn{1}{l|}{HUKB}          & \multicolumn{1}{l|}{HUKB1}             & \multicolumn{1}{c|}{0.6725}      & \multicolumn{1}{c|}{0.6279}     & 0.7079     \\ \hline
\multicolumn{1}{|l|}{5}           & \multicolumn{1}{l|}{LLNTU}         & \multicolumn{1}{l|}{LLNTUgigo}         & \multicolumn{1}{c|}{0.6535}      & \multicolumn{1}{c|}{0.7327}     & 0.6436     \\ \hline
\multicolumn{1}{|l|}{6}           & \multicolumn{1}{l|}{UA}            & \multicolumn{1}{l|}{TFIDF\_threshold2} & \multicolumn{1}{c|}{0.5642}      & \multicolumn{1}{c|}{0.6205}     & 0.5644     \\ \hline
\multicolumn{6}{|l|}{\textbf{Our results}}                                                                                                                                                        \\ \hline
\multicolumn{1}{|l|}{3}           & \multicolumn{1}{l|}{NOWJ}          & \multicolumn{1}{l|}{nowj.d-ensemble}   & \multicolumn{1}{c|}{0.7273}      & \multicolumn{1}{c|}{0.6823}     & 0.7673     \\ \hline
\end{tabular}
\label{tab: task3 method result}
\end{table}

\subsection{Task 4}
The problem raised on task 4 is tackled by utilizing the textual entailment head of the multi-task model trained on previous task. Although a multi-task approach can significantly raise the performance on the retrieval task, it does not improve the result of task 4 yet. Further research could be conducted on the utilization of this model for the textual entailment problem. Our team has submited 3 runs for the private test including the first run named \textbf{NOWJ.multi-v1-en} employed the English data for the training phase, the second one named \textbf{Multi-Task EN (NOWJ.multi-v1-en)} used Japanese data for the training phase and the final run (\textbf{NOWJ.multijp}) also utilizes Japanese data with a different inference strategy. The accuracy of our runs and other team's best results is shown at the Table \ref{tab: task4 method result}.

\begin{table}[ht]
\caption{Results on the private test of Task 4}
\centering
\begin{tabular}{|lllc|}
\hline
\multicolumn{1}{|c|}{\textbf{\#}} & \multicolumn{1}{c|}{\textbf{Team}} & \multicolumn{1}{c|}{\textbf{Best Run}} & \textbf{Accuracy} \\ \hline
\multicolumn{4}{|l|}{\textbf{Other team's result}}                                                                                  \\ \hline
\multicolumn{1}{|l|}{1}           & \multicolumn{1}{l|}{JNLP}          & \multicolumn{1}{l|}{JNLP3}             & 0.7822            \\ \hline
\multicolumn{1}{|l|}{2}           & \multicolumn{1}{l|}{TRLABS}        & \multicolumn{1}{l|}{TRLABS\_D}         & 0.7822            \\ \hline
\multicolumn{1}{|l|}{3}           & \multicolumn{1}{l|}{KIS}           & \multicolumn{1}{l|}{KIS2}              & 0.6931            \\ \hline
\multicolumn{1}{|l|}{4}           & \multicolumn{1}{l|}{UA}            & \multicolumn{1}{l|}{UA\_V2}            & 0.6634            \\ \hline
\multicolumn{1}{|l|}{5}           & \multicolumn{1}{l|}{AMHR}          & \multicolumn{1}{l|}{AMHR01}            & 0.6535            \\ \hline
\multicolumn{1}{|l|}{6}           & \multicolumn{1}{l|}{LLNTU}         & \multicolumn{1}{l|}{LLNTUdulcsL}       & 0.6238            \\ \hline
\multicolumn{1}{|l|}{7}           & \multicolumn{1}{l|}{CAPTAIN}       & \multicolumn{1}{l|}{CAPTAIN.gen}       & 0.5842            \\ \hline
\multicolumn{1}{|l|}{8}           & \multicolumn{1}{l|}{HUKB}          & \multicolumn{1}{l|}{HUKB1}             & 0.5545            \\ \hline
\multicolumn{4}{|l|}{\textbf{Our results}}                                                                                          \\ \hline
\multicolumn{1}{|l|}{9}           & \multicolumn{1}{l|}{NOWJ}          & \multicolumn{1}{l|}{multi-v1-jp}       & 0.5446            \\ \hline
\multicolumn{1}{|l|}{9}           & \multicolumn{1}{l|}{NOWJ}          & \multicolumn{1}{l|}{multijp}           & 0.5248            \\ \hline
\multicolumn{1}{|l|}{9}           & \multicolumn{1}{l|}{NOWJ}          & \multicolumn{1}{l|}{multi-v1-en}       & 0.4851            \\ \hline
\end{tabular}
\label{tab: task4 method result}
\end{table}

\section{Discussions}
In this section, we discuss the overall performance of our approaches across all tasks in the COLIEE 2023 Competition and the lessons learned from our participation.

\subsection{Task 1 and 2: Legal Case Retrieval and Entailment}

Our approach for Tasks 1 and 2, which relied on BERT and Longformer pre-training models, demonstrated promising results in legal case retrieval and entailment. The use of lexical matching for candidate case retrieval provided a good balance between computational cost and recall performance. However, there is room for improvement in the precision of our results. The use of BERT and Longformer in the matching phase aimed to address this issue by considering important paragraphs of base cases in a pairwise comparison with candidate cases. Although our team ranked third in both tasks, this suggests that further optimization of the models or the incorporation of additional features could improve our system's performance.

\subsection{Task 3: Statute Law Retrieval}

Our two-phase retrieval system for Task 3, which employed the BM25-ranking algorithm and a BERT-based model for re-ranking, proved effective at narrowing down candidate articles while maintaining a high recall score. The multi-task learning approach combined retrieval and textual entailment tasks to improve the precision of our results. Our ensemble method, based on voting, achieved the third rank among all teams. However, there is still room for improvement in terms of precision and recall balance. Future work could explore other ensemble techniques or refine the multi-task learning model to boost performance.

\subsection{Task 4: Legal Textual Entailment}

In Task 4, we utilized the textual entailment head of the multi-task model trained from Task 3. While this approach did not lead to significant improvements in performance compared to other teams' best results, it demonstrated the potential applicability of multi-task learning models in legal textual entailment problems. Further research is needed to better understand the limitations of our current model and to identify opportunities for improvement, such as refining the model architecture or incorporating additional features.

\subsection{Overall Lessons Learned}

Our participation in the COLIEE 2023 Competition has provided valuable insights into the challenges and opportunities in legal information processing. We have gained experience working with state-of-the-art machine learning models and innovative approaches, while also identifying areas for future research and improvement. Some key lessons learned include:

\begin{itemize}
    \item Understanding the importance of pre-processing and feature engineering in preparing data for legal information processing tasks.
    \item Recognizing the potential of pre-trained models like BERT and Longformer in addressing various legal information extraction and entailment challenges.
    \item Appreciating the effectiveness of multi-task learning models in combining different tasks to achieve better performance.
    \item Realizing the need for further research on domain-specific knowledge incorporation and model optimization to enhance our systems' performance in real-world legal scenarios.
\end{itemize}

\section{Conclusions}
In conclusion, our participation in the COLIEE 2023 Competition has allowed us to explore various state-of-the-art machine learning models and innovative approaches in the field of legal information processing. Although our team did not achieve the best results in all tasks, our performance in Tasks 1 and 2 with BERT and Longformer, as well as our two-phase retrieval system for Task 3 and the multi-task learning model for Task 4, demonstrated the potential of these methods in addressing real-world legal scenarios.

Our experience in this competition has provided valuable insights into the challenges and opportunities in legal information extraction and entailment. We believe that the methods and techniques we have employed can serve as a foundation for further research and improvements in this area. Future work could focus on refining the multi-task learning models, exploring other pre-training models, and incorporating domain-specific knowledge to enhance the performance of our systems.

Overall, our participation in the COLIEE 2023 Competition has been a fruitful learning experience, and we look forward to continuing our research on legal information processing and its applications in real-world scenarios.

\section*{Acknowledgement}
Hai-Long Nguyen was funded by the  Master, PhD Scholarship Programme of Vingroup Innovation Foundation (VINIF),  \\ code VINIF.2022.ThS.050.

\bibliographystyle{splncs04}
\bibliography{references.bib}
\end{document}